\documentclass[Afour,sagev,times]{sagej}

\usepackage{moreverb,url}

\usepackage[hidelinks=True,bookmarksopen,bookmarksnumbered,citecolor=black,urlcolor=black]{hyperref}

\newcommand\BibTeX{{\rmfamily B\kern-.05em \textsc{i\kern-.025em b}\kern-.08em
T\kern-.1667em\lower.7ex\hbox{E}\kern-.125emX}}

\def\volumeyear{2022}

\usepackage{amsmath} 
\numberwithin{equation}{section} 
\usepackage{bbold}

\usepackage{graphicx}
\graphicspath{ {images/} }
\usepackage{subcaption}
\usepackage[export]{adjustbox}
\usepackage{wrapfig}

\usepackage{caption}
\begin{document}

\runninghead{}

\title{Physically Meaningful Uncertainty Quantification in Probabilistic Wind Turbine Power Curve Models as a Damage Sensitive Feature}

\author{J.H. Mclean\affilnum{1}, M.R. Jones\affilnum{1}, B.J. O'Connell\affilnum{1}, A.E Maguire\affilnum{2}, T.J. Rogers\affilnum{1}}

\affiliation{\affilnum{1}Dynamics Research Group, Department of Mechanical Engineering, University of Sheffield, Mappin Street, Sheffield S1 3JD, UK\\
\affilnum{2}Vattenfall Research and Development, Holyrood Road, Edinburgh EH8 8AE, UK}

\corrauth{J.H. Mclean, Dynamics Research Group,
Sheffield S1 3JD, UK.}

\email{jmclean1@sheffield.ac.uk}

\begin{abstract}
A wind turbines' power curve is easily accessible damage sensitive data, and as such is a key part of structural health monitoring in wind turbines. Power curve models can be constructed in a number of ways, but the authors argue that probabilistic methods carry inherent benefits in this use case, such as uncertainty quantification and allowing uncertainty propagation analysis. Many probabilistic power curve models have a key limitation in that they are not physically meaningful – they return mean and uncertainty predictions outside of what is physically possible (the maximum and minimum power outputs of the wind turbine). This paper investigates the use of two bounded Gaussian Processes in order to produce physically meaningful probabilistic power curve models. The first model investigated was a warped heteroscedastic Gaussian process, and was found to be ineffective due to specific shortcomings of the Gaussian Process in relation to the warping function. The second model – an approximated Gaussian Process with a Beta likelihood was highly successful and demonstrated that a working bounded probabilistic model results in better predictive uncertainty than a corresponding unbounded one without meaningful loss in predictive accuracy. Such a bounded model thus offers increased accuracy for performance monitoring and increased operator confidence in the model due to guaranteed physical plausibility. 
\end{abstract}


\maketitle




\section{Introduction} \label{Introduction}

As global demand for offshore wind turbines continues to grow, so too does the need for cost effective methods of monitoring and maintaining these turbines. The nature of offshore environments means that cost and hazard of on-site monitoring and maintenance is high, and as such should be minimised where possible. Using autonomously collected data such as by the Supervisory 
Control and Data Acquisition  (SCADA) system (which come installed on most modern wind turbines) to detect faults and optimise maintenance is therefore desirable.

The power curve is commonly used in Structural Health Monitoring (SHM) as a damage sensitive feature, as seen in Papatheou et al. \cite{papatheou2015performance} and Gonzalez et al. \cite{gonzalez2019using}. The power curve describes the fundamental relationship of a wind turbine between wind speed and power output. The power curve can be modelled from SCADA data (which collect wind speed and power output data as a matter of course). New SCADA data that deviates from this power curve is a useful indication that something is wrong with the turbine and can be carried forward into further analysis and maintenance optimisation methods. Power curve modelling has benefits outside of SHM as well, such as use in financial return predictions, power output predictions or grid management. 

\begin{figure}[h]
    \centering
    \includegraphics[width=.75\linewidth]{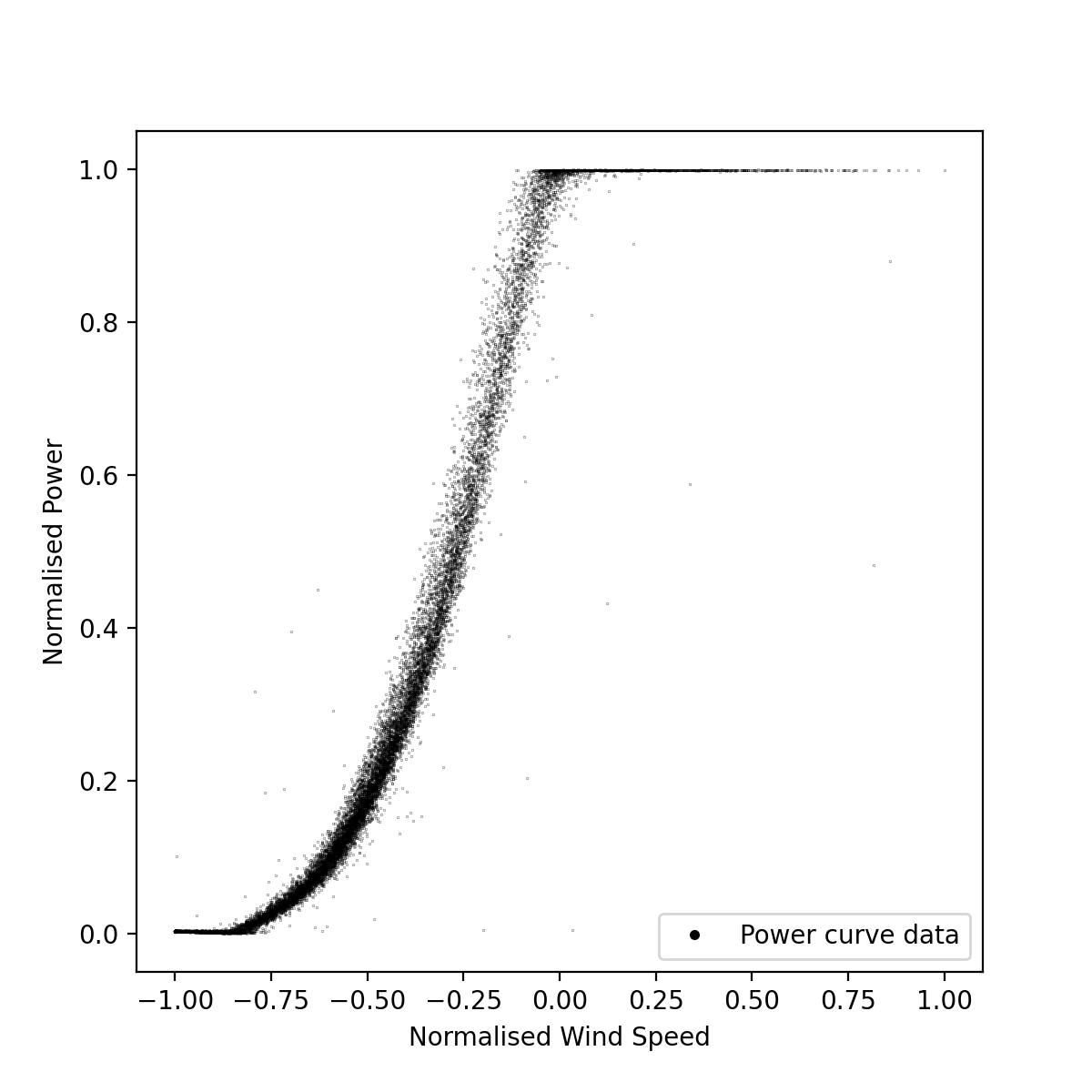}
    \caption{Example of a power curve}
    \label{fig:powerCurve}
\end{figure}

An example of a typical power curve is shown in \autoref{fig:powerCurve}. The power curve is bounded on two sides in the power dimension. The lower bound of the power curve occurs at zero power, where the wind does not contain enough energy to move the turbine. The upper bound is the maximum power that the turbine is designed to safely extract from the wind, also known as the rated power. If wind speed increases while the turbine is operating at rated power, it will regulate itself to maintain rated power before shutting down should speeds get too high. 

As seen in \autoref{fig:powerCurve}, the data that composes a power curve are inherently stochastic. The authors argue that the full richness of information in this noisy data cannot be truly represented by a deterministic approach, and should be modelled probabilistically for maximum utility. A probabilistic approach (specifically a Bayesian approach) provides value because it allows for the uncertainty of the predicted power outputs to be found. This additional information is useful because it can be used to understand and model uncertainty propagation on financial returns and structural health of the turbine, allowing for robust long term risk planning. The quantification of uncertainty also means that predictions are made with a quantified level of confidence, bolstering the decision making processes in relation to the turbine.

It is clear then that robust probabilistic models are beneficial to the industry, and provide tangible benefits to operators. Whilst probabilistic models are highly useful, it is important that the predictive distributions returned are physically meaningful if they are to be relied on for operational decision making. It is common for probabilistic models to assume (implicitly or explicitly) that the noise is Gaussian, and can be modelled as such even at the bounded limits of the power curve. This leads to predicted probability mass outside the realm of what is physically possible for the turbine to achieve, i.e. the models may suggest that there is some probability of producing more than a turbine's rated power. While the predictive density could simply be truncated this may not be statistically robust or representative of the true behaviour; it is clear that the  distribution below rated power is also not Gaussian. A robust method of modelling the uncertainty such that it remains physically meaningful is therefore desirable. 

This paper seeks to address this issue in probabilistic power curve modelling by investigating two bounded Gaussian Process (GP) models. The bounded models will be assessed against each other and a standard GP, which was generated as a performance baseline for comparison. The first bounding method uses a warping function to transform the data and output of the GP in order to satisfy the boundary conditions. This is coupled with a heteroscedastic\footnote{A heteroscedastic GP is a GP designed to model heteroscedastic noise, otherwise known as input dependant noise. This will be expanded upon in in the subsection ``Heteroscedastic noise".} GP, allowing the input dependent variance in the data to be captured as shown in Rogers et al. \cite{rogers2019bayesian}.

The second bounding method will replace the standard Gaussian likelihood in a GP with a Beta likelihood, again to satisfy boundary conditions. The Beta likelihood GP is approximated through quadrature and variational inference of a latent heteroscedastic GP. This method will be referred to as the Heteroscedastic Beta Process (HBP). The HBP method is beneficial because it models heteroscedasticity and the  non-Gaussian noise whilst allowing for the use of fast inference techniques, rather than relying on significantly slower approaches to approximate a non Gaussian posterior, such as Markov Chain Monte Carlo (MCMC) \cite{gamerman2006markov}. 

This paper is structured in the following manner. The section ``Related work" will provide a brief overview of the existing literature in wind turbine SHM and power curve modelling. The section ``Gaussian Process Regression" and its various subsections will cover the fundamental theory used to construct the bounded GP models. The section ``Case study" illustrates how the data was collected and used, outlines the metrics used to assess the success of the models and presents the performance results of the models. The subsections of section "Case study" provide a discussion on the results of their respective models. Finally, concluding remarks and a discussion on possible avenues are made in their respective sections. 

\subsection{Related work} \label{sec:Relatedwork}

The application of SHM in wind turbines is a well established field, and comprehensive reviews have been published previously \cite{martinez2016structural, marquez2012condition, ciang2008structural}. A wide variety of techniques exist and have been proposed for monitoring turbines, ranging from traditional Non Destructive Evaluation (NDE) approaches to entirely data based methods. NDE approaches are often limited to accessible parts of the turbine and require specialised sensors and equipment but can be highly effective at identifying faults in the turbine. Examples of NDE approaches for wind turbines include (but are not limited to) acoustic emissions \cite{dutton2003acoustic} and ultrasonic methods \cite{raivsutis2008ultrasonic} and thermal imaging \cite{dutton2004thermoelastic}. In an offshore environment, the periodic nature of NDE methods might be considered less attractive than the online and continuous monitoring offered by SHM methodologies. 

As mentioned in the introduction, most modern turbines come installed with a SCADA system. Using SCADA data for monitoring is a popular approach within the field, as it does not require additional equipment or expertise and allows for an online approach. This can be seen in Canizo et al.\cite{canizo2017real}, where a big data approach was used. Zaher et al. \cite{zaher2009online} applies a neural net to SCADA data to detect faults in the gearbox and generator. Yang et al. \cite{yang2013wind} uses SCADA data to detect blade and drive train faults. Finally, power curves using SCADA data are also commonly used data to identify faults in turbines \cite{papatheou2015performance, gonzalez2019using, uluyol2011power}.

Power curve modelling (whether related to SHM or not) is also an established area of research. A comprehensive literature review of power curve models was published by Lydia et al. \cite{lydia2014comprehensive}. Power curve models can be conceptually divided in numerous ways, but the most relevant division for this paper is that between deterministic and probabilistic models. 

Historically, power curve models have been simple polynomial models. Such models are, by default, deterministic but can be highly accurate. A wide variety of polynomial power curve models exist in the literature; ranging from a basic piece wise linear model \cite{khalfallah2007suggestions} to more complicated and commonly used models such as a cubic polynomial \cite{saint2020parametric}. High order polynomials have also been tested, such as the ninth order polynomial used in Raj et al. \cite{raj2011modeling}. However, higher order polynomial models can be prone to overfitting to the training data as discussed in Bishop \cite{bishop2006pattern}.

Modern approaches have steered away from classic polynomial models, and a variety of non-polynomial deterministic models have been utilised in the literature. Many of these power curve models are applications of classic machine learning algorithms; such as K Nearest Neighbour (K-NN) and random forest models \cite{kusiak2009line, schlechtingen2013using}, support vector machines \cite{ouyang2017modeling}, clustering center fuzzy logic modeling \cite{ustuntacs2008wind} and Copula modelling \cite{gill2011wind}. A non-linear regression model was used in Marčiukaitis et al. \cite{marvciukaitis2017non}. This is one of the few works in the deterministic power curve literature space which explicitly obeys physical limitations, however it is still a deterministic model and only returns confidence intervals. Neural nets are another popular approach to power curve models. Models can be fairly simple power curve models \cite{li2001using, panahi2015evaluation} or can be more complicated multi-input models such as in Pelletier et al. \cite{pelletier2016wind}. 

Probabilistic power curve models are comparatively rare, and often these models simply make no mention if the models or their returned distributions obey physical limitations. Probabilistic models vary widely in their approaches and chosen algorithms; from a probabilistic analytical solution \cite{jin2010uncertainty} to algorithms such as Monte Carlo and Fuzzy Clustering methods \cite{yan2019uncertainty} to a neural net used to estimate quantiles \cite{xu2021quantile}. Many probabilistic models assume Gaussianity in the power curve data, leading to physically impossible uncertainty densities. This can be seen in the heteroscedastic models in Rogers et al. \cite{rogers2019bayesian} and Wang et al. \cite{wang2018wind}, where Gaussianity was assumed and, although the heteroscedastic noise models reduced the issue, predictions still led to physically impossible distributions. The assumption of Gaussian noise is also made in Jin et al. \cite{jin2010uncertainty}, where it is only used to model the region between the bounds, lending credence to the idea that the Gaussian distribution cannot effectively model power output noise at the bounds of the power curve. Several physically meaningful probabilistic models exist in literature. The use of the Weibull distribution is common here, as seen in Ge et al. \cite{ge2020wind} and Yun et al. \cite{yun2021probabilistic}. The Weibull distribution is useful in this case because it is bound on its lower end, and as such can be used to reflect the bounded nature of the power curve in one direction. Finally, the Gaussian Process (GP) has also been used for probabilistic power curve modeling in literature. Papatheou et al. \cite{papatheou2015performance} uses a GP for outlier detection in power curves for SHM purposes. Rogers et al. \cite{rogers2019bayesian} uses a heteroscedastic GP is used to model the power curve, and Pandit et al. \cite{pandit2018scada, pandit2019comparative} also uses a GP for anomaly detection in power curves. 

Power curve modelling and SHM are active fields of research. As such, this brief review has not been able to cover all work completed in these field but has sought to highlight the relevant state-of-the-art and illustrate that physically meaningful probabilistic modelling is an important area that has seen some work but is comparatively under developed.

\section{Gaussian Process Regression} \label{Theory}

The Gaussian process \cite{rasmussen2003gaussian} is a probability distribution over the possible functions that fit a given set of points. The GP is designed to model regression problems of the form \(y = f(\Vec{x}) + \varepsilon\), where \(\varepsilon \sim \mathcal{N}(0, \sigma_n^2)\) is used to capture noise that may be present on the true function evaluations. $\Vec{x}$ is some vector of inputs and $\Vec{y}$ the target data with additive Gaussian noise $\varepsilon$ which has a variance $\sigma_n^2$. The GP is mathematically defined in \ref{gp_def}, where \(\mu(\Vec{x})\) is the mean function and \(k(\Vec{x},\Vec{x}^\prime)\) is the covariance function (otherwise known as the kernel). 


\begin{equation}\label{gp_def}
f \sim \mathcal{GP}(\mu(\Vec{x}),k(\Vec{x},\Vec{x}^\prime))
\end{equation}

GPs are a type of Bayesian inference, meaning that a defined prior is updated with observed data to produce a posterior distribution. The prior for the Gaussian process is the form of $\mu(\Vec{x})$ and $k(\Vec{x},\Vec{x}^\prime)$. The posterior can be computed through the use of Bayes theorem, by conditioning the the test data
\(\Vec{x}_*\) on the training data \(\Vec{x}\) through the joint Gaussian distribution (as given in \ref{multi_variate}). In \ref{multi_variate} the notation \(K_{\Vec{x}^*\Vec{x}}\) is used to write the covariance matrix between the test variables \(f(\Vec{x}^*)\) and training variables \(f(\Vec{x})\). Similar subscript notation is used for denoting covariance matrices between sets of training variable and test variables, e.g. \(K_{\Vec{x}\Vec{x}}\) for the covariance of the training data with itself.

\begin{gather} 
\begin{split} \label{multi_variate}
\begin{bmatrix} \Vec{y} \\ \Vec{y}_* \end{bmatrix}
\sim
\mathcal{N}
\begin{pmatrix}
\begin{bmatrix} m(\Vec{x})\\m(\Vec{x}_*) \end{bmatrix},
\begin{bmatrix}
K_{\Vec{x}\Vec{x}}+\sigma^2_nI & K_{\Vec{x}\Vec{x}_*}\\
K_{\Vec{x}_*\Vec{x}} & K_{\Vec{x}_*\Vec{x}_*}+\sigma^2_nI
\end{bmatrix}
\end{pmatrix}
\end{split}
\end{gather}

Assessing this distribution will yield the posterior predictive distribution of the GP, which shows that it is possible to predict new outputs \(\Vec{y}_*\) subject to new inputs \(\Vec{x}_*\) given the training inputs \(\Vec{x}\) and their respective outputs \(\Vec{y}\). The posterior predictive is given in \ref{post_pred}, where \(\mathbb{E}[\Vec{y}_*]\) and \(\mathbb{V}[\Vec{y}_*]\) are the mean and variance of the predicted distribution respectively. 

\begin{gather} 
\begin{split} \label{post_pred}
p(\Vec{y}_*|\Vec{x}_*,\Vec{y},\Vec{x}) \sim \mathcal{N}(\mathbb{E}[\Vec{y}_*],\mathbb{V}[\Vec{x}_*])
\end{split}
\\
\begin{split} \label{mean}
\mathbb{E}(\Vec{y}_*) = m(\Vec{x}_*)+K_{\Vec{x}_*\Vec{x}}(K_{\Vec{x}\Vec{x}}+\sigma^2_n \mathbb{I})^{-1}(\Vec{y}-m(\Vec{x}))\nonumber
\end{split}
\\ 
\begin{split} \label{variance}
\mathbb{V}(\Vec{y}_*) = K_{\Vec{x}_*\Vec{x}_*}-K_{\Vec{x}_*\Vec{x}}(K_{\Vec{x}\Vec{x}}+\sigma^2_n \mathbb{I})^{-1}K_{\Vec{x}\Vec{x}_*}+\sigma_n^2\mathbb{I}\nonumber
\end{split}
\end{gather}

\ref{gp_def} shows that the GP is fully defined by its mean and covariance functions. The mean is commonly fixed to zero in literature \cite{williams2006gaussian} and in practice (as is the case in this paper). The covariance function specifies the family of distributions that may have generated the observed data. Each kernel enforces different types of behavior on the corresponding function draws, with the ability to combine kernels allowing for a number of different types of structure to be embedded into the GP prior. A popular kernel for GPs is the squared exponential kernel, given in \ref{skernel}, where \(\sigma^2\) is the signal variance and \(l\) is the lengthscale of the kernel, these two values are the \emph{hyperparameters} of the kernel.


\begin{equation}\label{skernel}
k(\Vec{x},\Vec{x}')=\sigma^2\exp\left({-\frac{(\Vec{x}-\Vec{x}')^2}{2l^2}}\right)
\end{equation}

Signal variance and length scales are examples of hyperparameters; these control the shape of the family of functions defined by the kernel (and by extension the GP). They are not known \emph{a priori}, and are learnt by minimising the negative log of the marginal likelihood of the model (\ref{NLML}), which solves a Type-II maximum likelihood problem such that \(\hat{\theta} = \underset{\theta}{\text{argmin}} \left\{-\log p(\Vec{y}|\Vec{x},\theta)\right\}\).

\begin{equation} \label{NLML}
\begin{split}
\log{p(\Vec{y}|\Vec{x},\theta)} = &-\frac{1}{2}\Vec{y}^T[K+\sigma^2_nI]^{-1}_{\Vec{y}}\Vec{y} \\ &-\frac{1}{2}\log{[K+\sigma^2_nI]}-\frac{n}{2}\log{2\pi}
\end{split}
\end{equation}

\subsection{Heteroscedastic noise}\label{Heteroscedastic noise}

Heteroscedastic noise is noise that varies as a function of a function's input. The power curve data in \autoref{fig:powerCurve} is an excellent example of this; it is clear that the data is much more noisy in the centre of the power curve than towards its limits - the input (the wind speed) is correlated to the amount of noise, i.e. level of uncertainty. The standard (homoscedastic) GP assumes that noise and hence uncertainty is constant across the full input range, and as such cannot model the way the noise changes with wind speed. In order to model the heteroscedastic noise and return a more useful probabilistic power curve, a heteroscedastic GP model can be used, as was done in Rogers et al. \cite{rogers2019bayesian}. 

The particular heteroscedastic model adopted in this work (distinct from Rogers et al. \cite{rogers2019bayesian}) uses two latent GPs to learn the location (\ref{locfx}) and scale (\ref{scalefx}) of the distributions required, first shown in Lázaro-Gredilla et al. \cite{lazaro2011variational}. This is done by connecting a Multi-output kernel to a heteroscedastic likelihood. This maps the latent GPs onto a single function. The heteroscedastic GP model can be described as:
\begin{subequations}
\begin{equation} \label{locfx}
        f_1(\Vec{x}) \sim \mathcal{GP}(\mu(\Vec{x}),k_1(\Vec{x},\Vec{x}^\prime))
\end{equation}
\begin{equation} \label{scalefx}
        f_2(\Vec{x}) \sim \mathcal{GP}(\mu(\Vec{x}),k_2(\Vec{x},\Vec{x}^\prime))
\end{equation}
\begin{equation}
    \text{m}(\Vec{x}) = f_1(\Vec{x})
\end{equation}
\begin{equation} \label{g_equation}
    \text{s}(\Vec{x}) = \text{g}(f_2(\Vec{x}))
\end{equation}
\begin{equation}
y(\Vec{x}) \sim \mathcal{N}(m(\Vec{x}), s(\Vec{x}))  
\end{equation}
\label{eq:hetero_GP}
\end{subequations}

The transform \(g(\cdot)\) in \ref{g_equation} in this case is an exponential transform, in order to keep the function positive, since it models the variance of the noise which must be greater than zero. In the model defined in \ref{eq:hetero_GP}, the two latent GPs have been assumed independent. However, this does not necessarily have to be true - the latent GPs can be modelled as dependent should the need arise by linking \ref{locfx} and \ref{scalefx} through a multi-output GP, e.g. Alvarez et al. \cite{alvarez2011computationally}. 

\subsection{Warping functions}\label{warp_theory}

\begin{figure*}[h]
     \centering
     \begin{subfigure}[t]{0.3\textwidth}
         \centering
         \includegraphics[width=\textwidth]{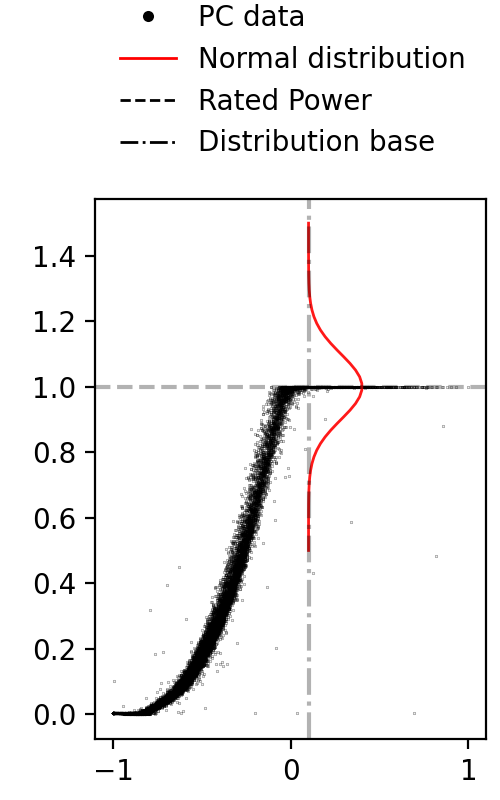}
         \caption{\centering Gaussian uncertainty in bounded space}
         \label{fig:WE_1}
     \end{subfigure}
     \begin{subfigure}[t]{0.3\textwidth}
         \centering
         \includegraphics[width=\textwidth]{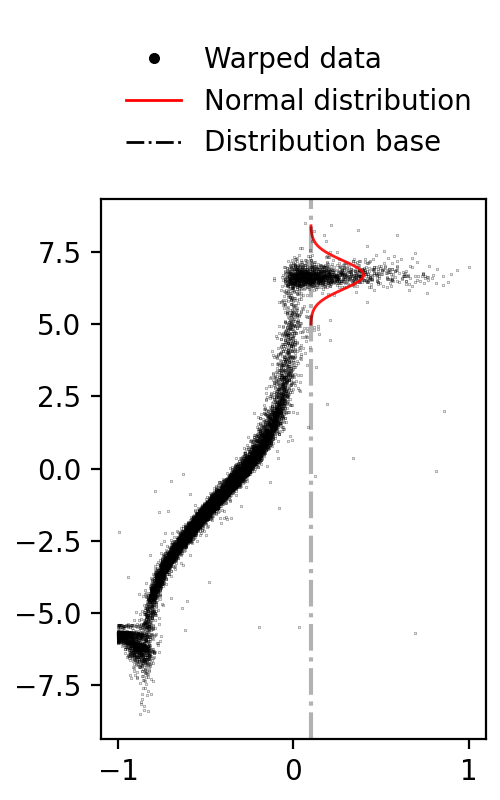}
         \caption{\centering Gaussian uncertainty in unbounded space}
         \label{fig:WE_2}
     \end{subfigure}
          \begin{subfigure}[t]{0.3\textwidth}
         \centering
         \includegraphics[width=\textwidth]{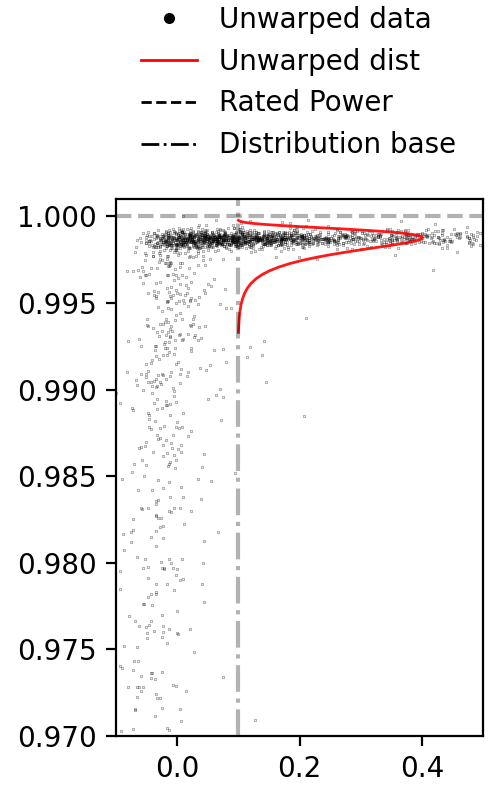}
         \caption{\centering Unwarped distribution in bounded space}
         \label{fig:WE_3}
     \end{subfigure}

        \caption{Illustration of the warping process - the normal distributions are for illustrative purposes only}
        \label{fig:The_bigWE}
\end{figure*}

The first bounded model investigated in this paper is the warping function. A visual illustration of how the warping function works is shown in \autoref{fig:The_bigWE}. \autoref{fig:WE_1} is an illustrative example of how a simple normal distribution modelling uncertainty at rated power has a large portion of probability mass outside of what is physically possible. 

The warped bounding method starts with transforming the power data using the logit function \cite{swiler2020survey}, shown in \ref{logit}. Conceptually, the warped GP moves the data from the bounded space to an unbounded one by means of a nonlinear transformation. The warped power curve is shown in \autoref{fig:WE_2}. The logit function moves the power curve from a bounded space (\(\text{p} \in (0,1)\)) in the power dimension to an unbounded space (\(\text{p} \in (-\infty,\infty)\)) in the power dimension. The GP is then applied to the warped power curve, as in this unbounded space the bounds on the power curve no longer apply, so the assumption of Gaussianity holds. 

\begin{equation}\label{logit}
        \text{logit}(p) = \ln (\frac{p}{1-p}) \text{ for p} \in (0,1)
\end{equation}

After fitting, the data and the returned uncertainty distribution are unwarped using the inverse of the logit function, resulting in the bounded data and distribution shown in \autoref{fig:WE_3}. An important note on this method is that it is difficult to evaluate the unwarped uncertainty, as no closed form solution exists for it; the density shown in \autoref{fig:WE_3} is presented for visualisation but is not exact. Therefore, it is simpler to work with and evaluate the model directly in the warped space. To illustrate how the uncertainty looks in the unwarped space for the entire power curve, a confidence bound can be created in the warped space at \(\pm 2\sigma\) and unwarped with the predictive mean. 

\subsection{Non Gaussian likelihoods}\label{Non-Gaussian likelihoods}

A typical GP model is formulated using a Gaussian likelihood. This assumes that the data is drawn from a normal distribution, i.e. f(x) has noise \(\varepsilon\) added and \(\varepsilon \sim \mathcal{N}(0,\sigma_n^2)\). This results in a normally distributed posterior function, as shown in \autoref{fig:WE_1}. Without any modifications, this is not a suitable model for probabilistic power curve modelling. Instead of warping the output space, the likelihood can be changed to a non Gaussian likelihood - one that better represents the distribution of the observed data for problem at hand. 

The likelihood chosen here is the Beta likelihood, imposing the assumption that the data is drawn from a Beta distribution. The Beta distribution is a continuous probability distribution in the interval [0,1]. It is defined in \ref{beta_dist} where \(\alpha\) and \(\beta\) are the shape parameters of the distribution. The probability density function is given in \ref{beta_pdf}, where \(B(\alpha, \beta)=\Gamma(\alpha)\Gamma(\beta)/\Gamma(\alpha+\beta)\) and \(\Gamma\) is the Gamma function. The use of the Beta likelihood is a bounding method because the GP with a Beta likelihood would imply that data were generated from Beta distributions, which are inherently bound between [0,1].
\begin{gather} 
    \begin{split}\label{beta_dist}
        y \sim \text{Beta}(\alpha, \beta)
    \end{split}
    \\
    \begin{split}\label{beta_pdf}
       p(y) = y^{\alpha-1}\frac{(1-y)^{\beta-1}}{B(\alpha,\beta)} 
    \end{split}
\end{gather}
Using a non Gaussian likelihood in the GP (like the Beta likelihood) results in an analytically intractable posterior \cite{jensen2013bounded}. The posterior must instead be approximated, numerous approximations exist for this problem. A common method (used in the popular GP packages GPflow \cite{GPflow2017} and GPML \cite{10.5555/1756006.1953029}) to approximate the posterior is MCMC \cite{gamerman2006markov} sampling. However, this can require a large amount of samples to approximate the posterior well, and can be very slow. 


Instead of using MCMC, this paper approximates the Beta likelihood using variational inference and quadrature of a heteroscedastic GP. The model proceeds by introducing two latent GPs to calculate \(\alpha\) and \(\beta\) as functions of the input, instead of the mean and variance as used in the normal heteroscedastic GP in \autoref{Non-Gaussian likelihoods}. This is expressed mathematically below, where \(t(\cdot)\) represents an exponential transform to keep the functions over \(\alpha\) and \(\beta\) positive.

\begin{gather}
    \begin{split}\label{f1_f2}
        [\Vec{f_1},\Vec{f_2}]=GP(m(\Vec{x}),k(\Vec{x},\Vec{x}'))
    \end{split}
    \\
    \begin{split}\label{g1}
        \Vec{g_1}=t(\Vec{f_1})
    \end{split}
    \\
    \begin{split}\label{g2}
        \Vec{g_2}=t(\Vec{f_2})
    \end{split}
    \\
    \begin{split}\label{betag1g2}
        \Vec{y} \sim \text{Beta}(\Vec{g_1},\Vec{g_2})
    \end{split}
\end{gather}

\begin{figure*}[h!]
    \centering
    \includegraphics[width=.8\textwidth]{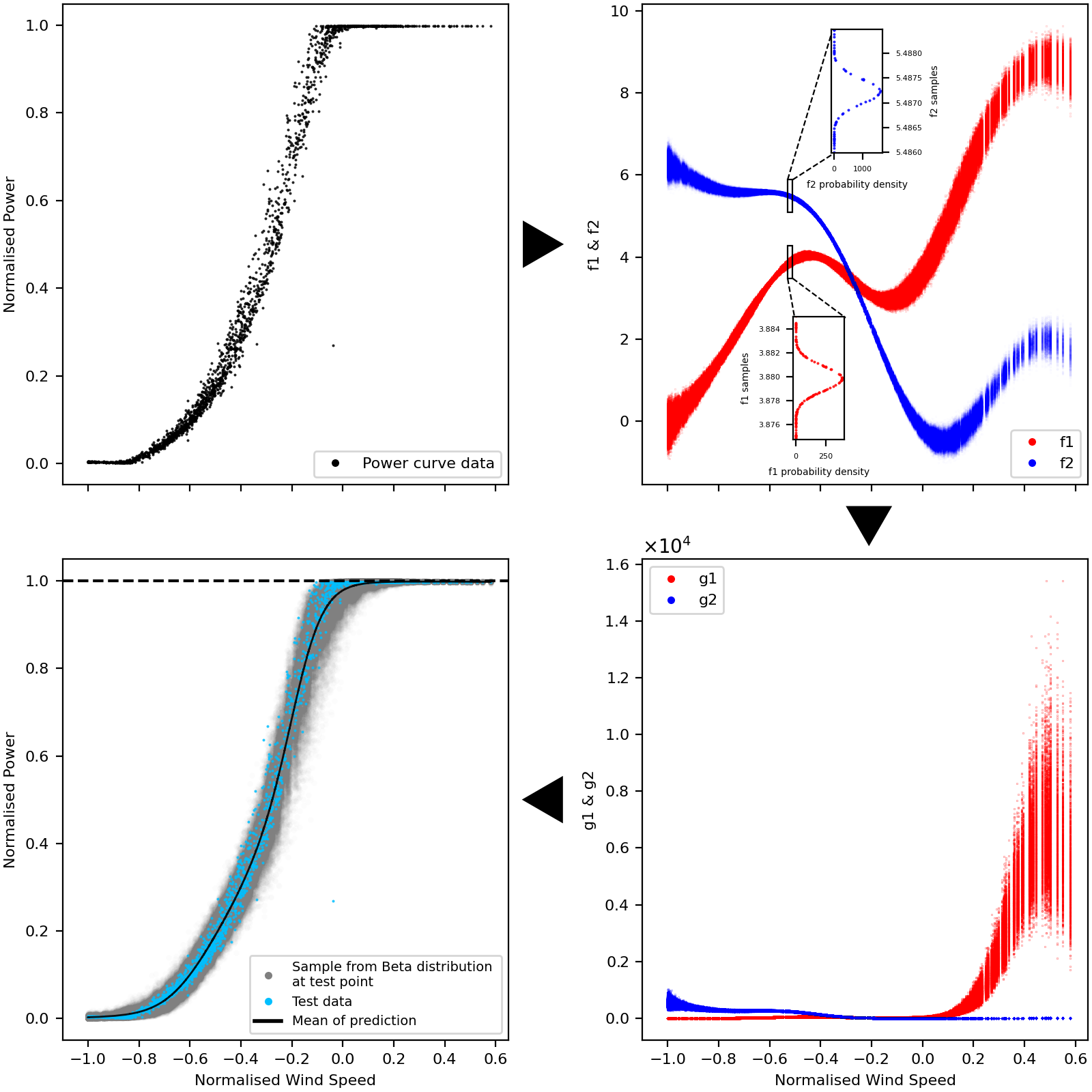}
    \caption{Flow chart of the Heteroscedastic Beta Process}
    \label{fig:BetaExplanation}
\end{figure*}

A flow diagram of this process is shown in \autoref{fig:BetaExplanation}. \autoref{fig:BetaExplanation} illustrates how \(f_1\) and \(f_2\) are calculated from the power curve data using the heteroscedastic GP, both \(f_1\) and \(f_2\) are distributed normally for each input. These are then transformed exponentially to create \(g_1\) and \(g_2\) which are approximations of \(\alpha\) and \(\beta\). Using total variance and mean, the distributions of \(g_1\) and \(g_2\) are used to find a final \(\alpha\) and \(\beta\) value for each wind speed input. These are subsequently used to create the final Beta distributions at each input shown in the final panel. 

This approach is useful because it allows for fast posterior approximation with established methods, in this case the Stochastic Variational Inference approximation (SVGP) \cite{hensman2013gaussian}. This is further explained in the following section.

A further benefit to this approach is that the heteroscedasticity of the model is not lost, as \(\alpha\) and \(\beta\) vary with inputs and as such capture the varying uncertainty with inputs as one would expect, alongside capturing the progression of the mean function.

\subsection{Approximations and Quadrature solutions} \label{Approximations and Quadrature solutions}

When considering the large heteroscedastic models used here, two problems arise. The first is that solutions for heteroscedastic and non-Gaussian models do not, in general, exist in closed form, and must be approximated to reach a solution. The second is that GPs are computationally inefficient when used on large datasets. This is due to their reliance on a matrix inversion when calculating the posterior and the marginal likelihood, which has time complexity \(\mathcal{O}(N^3)\) and therefore scales poorly with increasing numbers of training data points. The SCADA dataset being used in this paper is quite large so calculating the entire posterior is impractical and would require an excessive amount of computational resource and time. 

This paper uses a stochastic variational inference approach \cite{hensman2013gaussian} in order to address these problems. Similarly to other sparse variational inference methods, the method applied in this work uses a set of inducing points \(\Vec{u}\) to summarise the model. The approximated posterior is given in \ref{aprox_post}. 
\begin{equation} \label{aprox_post}
q(f, u) = p(f, u)\phi(u)
\end{equation}

The approximate posterior is found by maximising the bound on the marginal likelihood. This bound is known as the Evidence Lower Bound (ELBO) and is used instead of the marginal likelihood to reduce computational cost. The global stochastic variational approach \cite{hensman2013gaussian} differs from other variational inference bounds in that it allows for the inducing points to be global points representing the entire GP despite being calculated locally - thus allowing faster computation. 

A formulation of the evidence lower bound (ELBO) of a heteroscedastic GP\cite{saul2016chained} is shown in \ref{ELBO}, where \(\int q(f_i) q(g_i) \log p (y_i|f_i, g_i) df_i dg_i\) is the likelihood term. The terms \(q(f_i)\) and \(q(f_i)\) are the marginal variational distributions of the mean and variance latent GPs respectively, and \(\log p(y_i|f_i, g_i)\) is the conditional probability term. The  model can then be optimised from this bound using stochastic gradient descent. 

\begin{align}
    \begin{split} \label{ELBO}
        \log p(\Vec{y}) \geq & \sum^n_{i=1} \int q(f_i) q(g_i) \log p (y_i|f_i, g_i) df_i dg_i \\
        & -\text{KL}(q(u_f)||p(u_f)) -\text{KL}(q(u_g)||p(u_g))
    \end{split}
\end{align}

The likelihood term is clearly an $n$ term integration over the marginal variational distributions and the conditional probability. As long as these terms are all Gaussian (such as in the heteroscedastic GP model), this integration has an analytical solution. However, the conditional term \(\log p (y_i|f_i, g_i)\) is not Gaussian in the HBP model, and as such the likelihood term does not have an analytical solution. It is therefore approximated with quadrature, as shown in Saul et al. \cite{saul2016chained}.

The use of sparse variational inference allows the GP to be computed with time complexity \(\mathcal{O}(N M^2)\) instead of \(\mathcal{O}(N^3)\), where $M$ is the number of inducing points. This solves the computational complexity problem, as it now requires significantly less time to train the heteroscedastic models. As mentioned earlier, the approximated posterior is found by maximising the ELBO. This means that the heteroscedastic methods can be used, as they can be approximated efficiently.

\section{Case study} \label{Modelling}

The data used in this paper comes from the supervisory control and data acquisition (SCADA) system of an operating wind turbine. The SCADA data is the 10 minute average of power output and wind speed (as taken on the nacelle) is comprised of 50,000 readings. The measured values of wind and power have been obscured by scaling for confidentiality reasons \footnote{Due to the confidentiality of the data, it is not possible to release the data used in this case study, however the code used to generate the results is made available at https://releasedonacceptance.}.

The dataset was split into three equal parts: a training dataset, a testing dataset and a validation set. The training data is used to optimise the hyperparameters of the GP. The predictive performance of the trained models are evaluated on the test data, which they have not seen before. The validation set is put aside should further validation of predictive performance be required.

After the initial scaling to anonymity, the power output data was again normalised between 0 and 1 in order to allow for the Beta likelihood and logit warping approaches to work. Neither the original scaling or the subsequent normalisation impacted the relationship between power output and wind speed as both were linear transforms. The training data was then cleaned of severe outliers, for example known curtailments.

Two performance metrics will be used in order to compare the success of the various models in modelling the power curve. The first is Normalised Mean Squared Error (NMSE), shown in \ref{NMSE}, which assesses how close to the mean of the predictive distribution the test value was. The NMSE can be thought of (and functions similarly to) percentage error.

\begin{equation}\label{NMSE}
    \text{NMSE} = \frac{100}{N\sigma^2}\sqrt{(y-\hat{y})^T(y-\hat{y})}
\end{equation}

The second metric being used is the joint log predictive likelihood which is computed as the sum of the predictive log likelihood at every test value under the posterior.  This is an especially important metric for this paper as it is a measurement of how good the uncertainty prediction of a specific model was at capturing the test data and not solely the performance of the mean estimate. The larger the joint log likelihood, the better the model has captured the true distribution of the test data and the more successful the model. Interpretation of the absolute values of the log likelihood is difficult, however, comparatively models which better capture the distribution of the data will have higher log likelihoods. 

To facilitate the discussion of results, the NMSE and joint log likelihood of each model is shown in \autoref{tab:Case study results}, where WH GP is the Warped Heteroscedastic GP. The results of each model will be further discussed in their respective following subsections.

\begin{table}[h!]
\begin{center}
\begin{tabular}{| c | c | c |}
\hline
\textbf{Model} & \textbf{NMSE} & \textbf{Joint log likelihood}\\
\hline
Standard GP & \textbf{0.24} & 32312 \\ \hline
WH GP & 5.4 & -12469 \\ \hline
HBP & 0.53 & \textbf{68866} \\ \hline
\end{tabular}
\vspace{1em}
\caption{\label{tab:Case study results}Case study results}
\end{center}
\end{table}

It will be argued by the authors in the remainder of this section that the HBP best captures the physical behaviour and uncertainty associated with the power curve of an operating wind turbine. All the models presented in this paper were written in Python and run using GPflow \cite{GPflow2017}.

\subsection{Simple GP Regression} \label{Simple GP Regression}

A standard (homoscedastic sparse) GP model was trained over the power curve to act as a benchmark to the two bounded models being tested. The standard GP used an additive composite kernel compromised of a Mat\'ern 3/2 kernel and a Linear kernel. This combination was used to capture the upward trend of the data.

\begin{figure*}[h]
     \centering
     \begin{subfigure}[t]{0.49\textwidth}
         \centering
         \includegraphics[width=\textwidth]{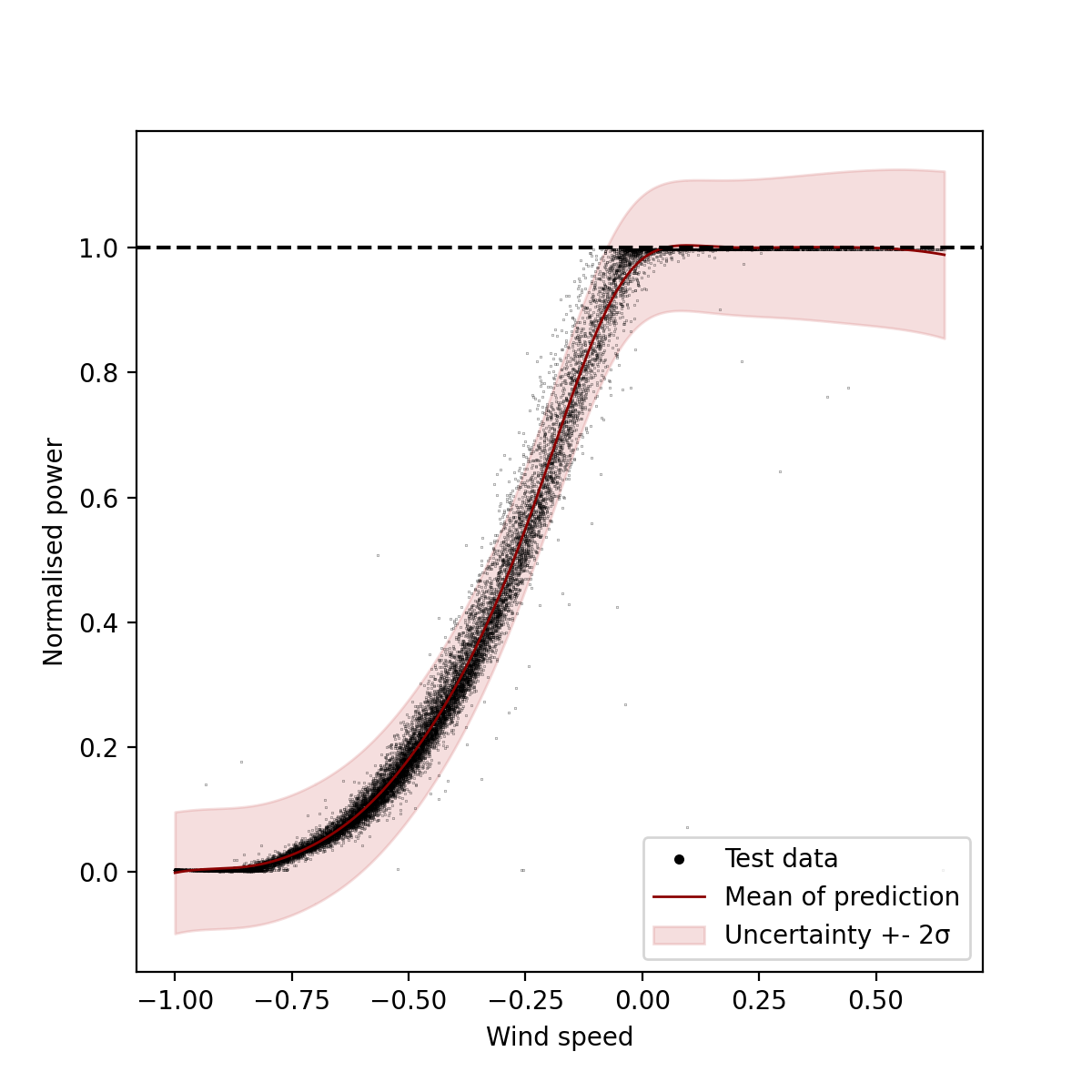}
         \caption{\centering Power curve model of standard GP}
         \label{fig:VanillaGP}
     \end{subfigure}
     \begin{subfigure}[t]{0.49\textwidth}
         \centering
         \includegraphics[width=\textwidth]{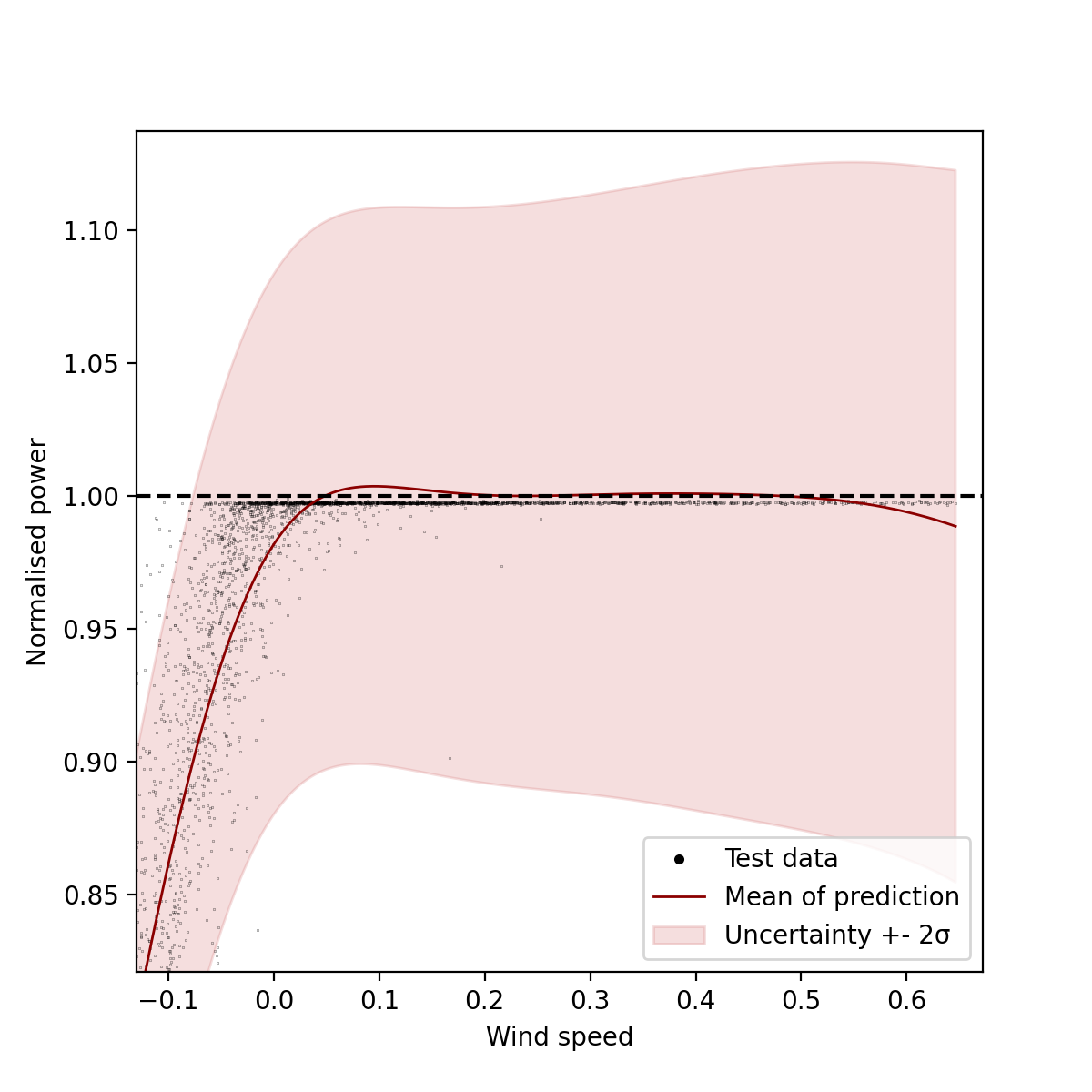}
         \caption{\centering Rated power model performance of Standard GP}
         \label{fig:VGP_ZOOM}
     \end{subfigure}
        \caption{Standard GP result}
        \label{fig:BIG_VanillaGP}
\end{figure*}

In \autoref{fig:VanillaGP} the results of the standard GP model can be seen. The model seemingly performs well in its mean predictions with an NMSE of 0.24. It also shows a log likelihood of 32312, less than half of that seen for the HBP, indicating worse performance in the uncertainty quantification. \autoref{fig:VGP_ZOOM} shows a magnified portion of the same model at rated power in order to examine how successful the model was at quantifying uncertainty at rated power. The standard GP model captured the shape of the power curve very well on average across the range of wind speeds, and the NMSE reflects this. The uncertainty prediction is comparatively poor; the uncertainty distribution is underestimated away from the bounds and overestimated close to them. This can be seen in that some data is not covered by the distribution in the middle of the power curve, and that the distributions at rated and zero power overestimate the variance. 

While the GP has an excellent NMSE and good joint log likelihood, it is clear from visual examination of \autoref{fig:VGP_ZOOM} that the predicted mean and uncertainty do not obey physical limits and overestimate possible values of the rated power. Perhaps the point of most concern here, is that depsite the excellent NMSE score, there are still predictions above the rated power output. This output is then not a particularly useful model if carried forward into further analysis, as the errors in the uncertainty and mean estimation could mask damage in the turbine or suggest that the financial performance of the wind could exceed what is physically possible. 


Finally, it is clear that the GP struggles to correctly predict the transition to rated power. This could be because continuous function models such as the GP can struggle with abrupt transitions such as this one. There are ways of addressing this shortcoming of GPS, such as using Bayesian committees to model the transition to rated power \cite{rogers2019bayesian}.

\subsection{Heteroscedastic warped Gaussian process} \label{Heteroscedastic warp function}

\begin{figure*}[h!]
     \centering
     \begin{subfigure}[t]{0.3\textwidth}
         \centering
         \includegraphics[width=\textwidth]{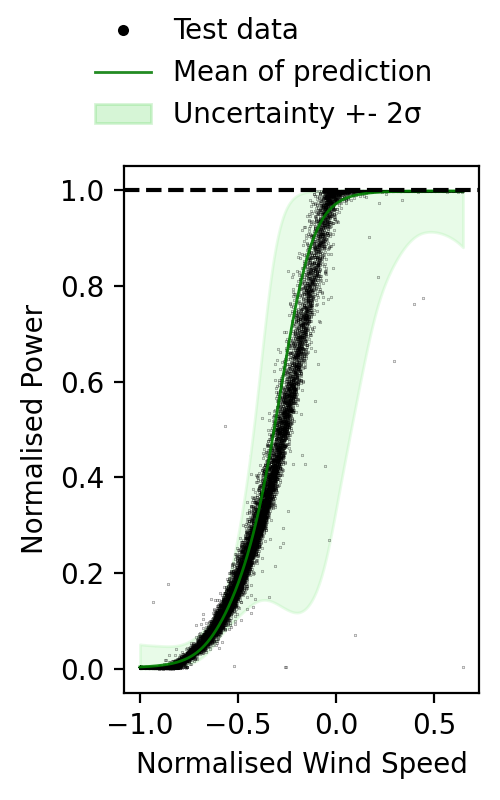}
         \caption{\centering Power curve model of warped Heteroscedastic GP in the original data space}
         \label{fig:WarpGP}
     \end{subfigure}
     \begin{subfigure}[t]{0.3\textwidth}
         \centering
         \includegraphics[width=\textwidth]{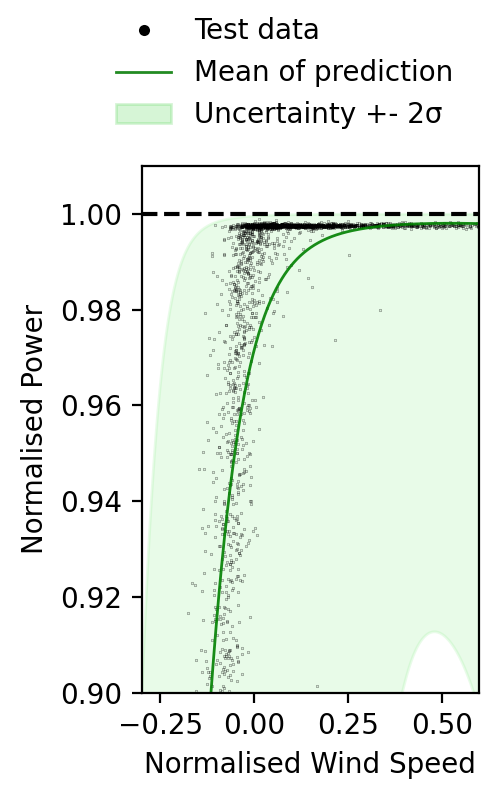}
         \caption{\centering Magnification of rater power}
         \label{fig:WARP_ZOOM}
     \end{subfigure}
          \begin{subfigure}[t]{0.3\textwidth}
         \centering
         \includegraphics[width=\textwidth]{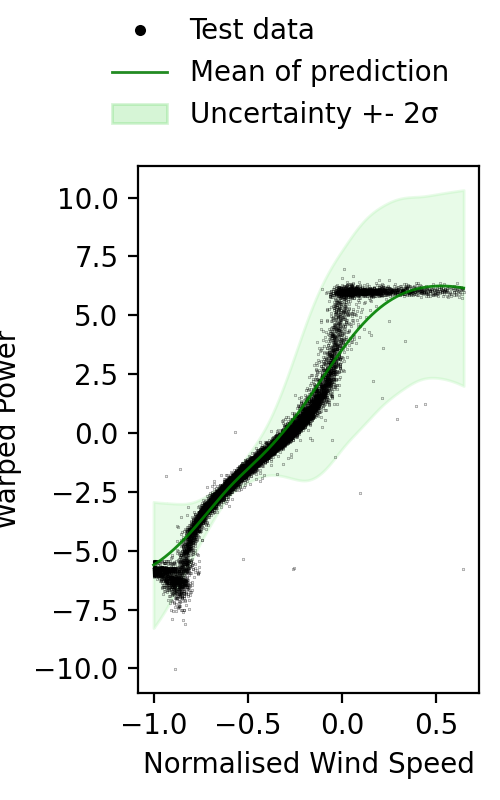}
         \caption{\centering Heteroscedastic GP in the warped space}
         \label{fig:warped_space}
     \end{subfigure}
        \caption{Heteroscedastic GP results}
        \label{fig:BIG_WarpGP}
\end{figure*}

The prediction of the power curve with the warped GP is shown in \autoref{fig:WarpGP}, and a magnified image of the model at rated power is shown in \autoref{fig:WARP_ZOOM}. The performance metrics for this model were comparatively very poor, with an NMSE of 5.4 and a log likelihood of -12469. The warping approach has failed to correctly capture either the mean behaviour or the uncertainty in the power curve.

The performance metrics clearly show that the warped GP has performed poorly. In order to understand why, the fitted GP must be examined in the warped space (shown in \autoref{fig:warped_space}), as this is where the GP interacts with the data being modelled. It is clear from \autoref{fig:warped_space} that the GP has failed to map the warped function correctly. This is likely because warping the power curve has resulted in a data shape that is too noisy and discontinuous for the GP with a Gaussian likelihood to model well. 

The inherent noise of the power curve in the power axis was exacerbated during the warping process, resulting in large vertical noise distributions that the GP cannot capture except by increasing the variance. This would confuse the training of the GP and is why such large uncertainty distributions are returned in \autoref{fig:WarpGP} and \autoref{fig:warped_space}. The discontinuity of the warped shape is also a problem because the GP is a continuous function and struggles to model abrupt changes as discussed in the previous section and in Cornford et al. \cite{cornford1998adding}. This results in the GP smoothing over the abrupt changes in the model as seen in \autoref{fig:warped_space}. It appears that this challenge in modelling the discontinuous transition is amplified when projecting the data through the warping function.

Despite being an intuitive approach, the heteroscedastic warped GP is clearly not that useful in assessing either the predictive mean or the uncertainty. The issues with this model lie in the warped space, where the data shape becomes too noisy and discontinuous. These problems could be addressed in several ways such as with further cleaning in the warped space, with a Bayesian committee \cite{rogers2019bayesian} or an alternative warping function (although it is not clear what that warping function may be).


\subsection{Beta Likelihood} \label{Beta Likelihood}

\begin{figure*}[h!]
     \centering
     \begin{subfigure}[t]{0.3\textwidth}
         \centering
         \includegraphics[width=\textwidth]{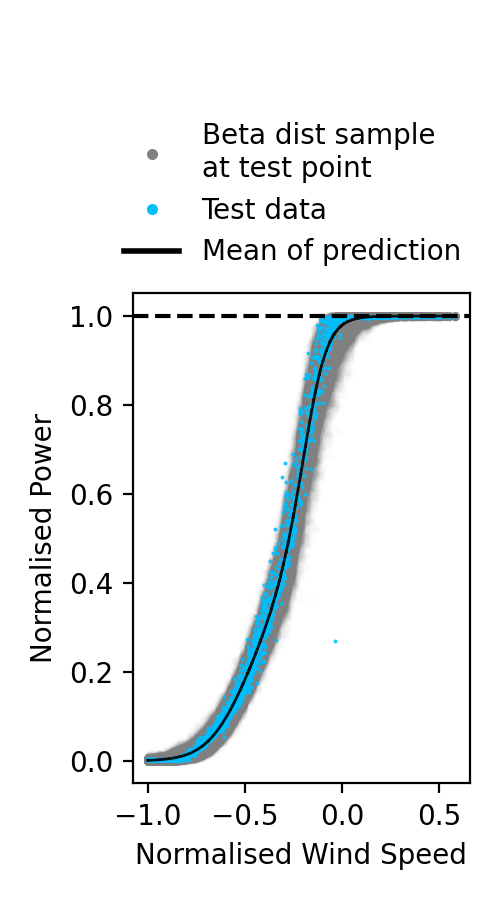}
         \caption{\centering HBP power curve}
         \label{fig:Beta_GP}
     \end{subfigure}
     \begin{subfigure}[t]{0.3\textwidth}
         \centering
         \includegraphics[width=\textwidth]{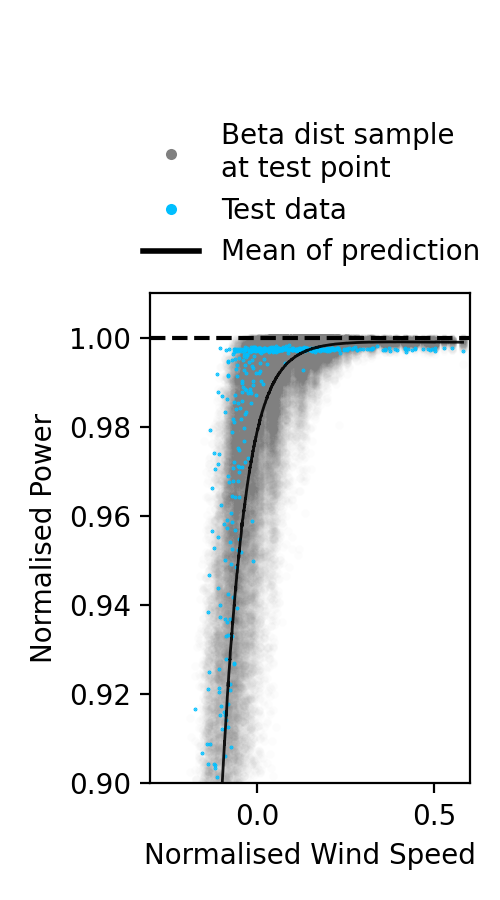}
         \caption{\centering Magnification of rater power}
         \label{fig:BGP_ZOOM}
     \end{subfigure}
          \begin{subfigure}[t]{0.3\textwidth}
         \centering
         \includegraphics[width=\textwidth]{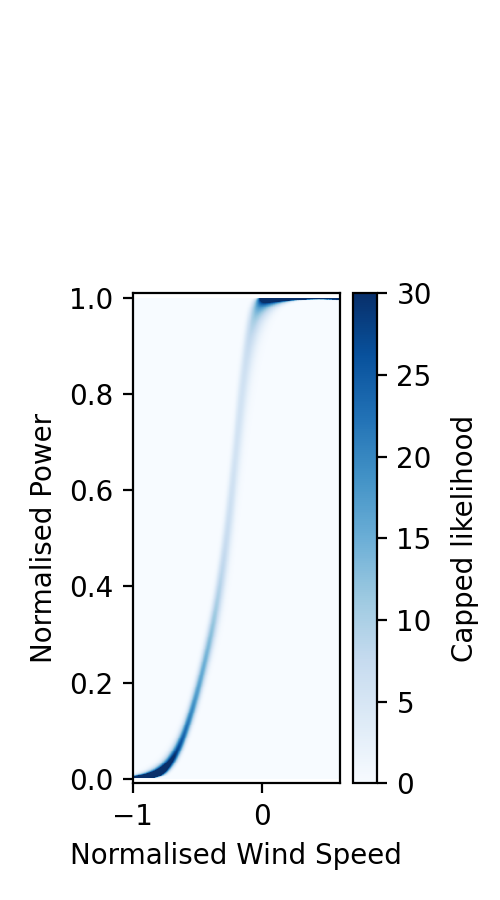}
         \caption{\centering Capped heat map of the predictive uncertainty of the HBP}
         \label{fig:heatmap}
     \end{subfigure}
        \caption{HBP results}
        \label{fig:BIG_BETA_GP}
\end{figure*}

The results from the HBP model can be seen in \autoref{fig:Beta_GP}. A magnification of the model at rated power is shown in \autoref{fig:BGP_ZOOM}. \autoref{fig:heatmap} is a heat map of the predictive uncertainty over the test data in order to illustrate that the uncertainty is recoverable as a closed form solution. The heat map is capped at 30 due to the likelihood being so high towards the bounds.

The NMSE for this model is 0.53, while the log likelihood is 68866. These are excellent results. The log likelihood is the highest of all the models tested, and shows that the bounded distribution best captures the power curve data. The NMSE is higher than for the standard GP, however the difference (0.53 vs 0.24) is small enough that these results can be considered comparable. It is worth noting again the previous discussion regarding the failure of the standard GP model to obey the physical bounds of the system in both the mean and the variance despite its lower NMSE.  

Choosing between the standard GP and HBP will depend on the desired use case. When uncertainty matters, the HBP is clearly a superior model. If only the mean prediction matters then the standard GP could be considered the better choice, however it is worth considering that the lower NMSE of the standard GP is misleading; the NMSE is simply an average of performance over the curve, and will not penalise physically implausible predictions. As seen in \autoref{fig:VGP_ZOOM} the predictive mean of the standard GP does exceed rated power, and as such is physically implausible. Blindly using the unbounded model could have serious consequences should its predictions be taken forward without further handling - such as over-predicting financial returns or masking turbine damage.

As seen in \autoref{fig:Beta_GP} the model has captured the shape of the power curve well. Similarly to the other models, this model also struggles to accurately fit the transition to rated power. This is marginally less pronounced as the use of Beta distributions allow the mean to be ``squashed" into the transition better than a normal distribution would allow. 

It is clear from the performance metrics that the HBP model is successful. It presents significant utility in that a full bounded predictive uncertainty distribution is returned. This uncertainty distribution also captures the data better than the standard GP and warped model. This has significant benefits as an improved power curve model for applications outside of SHM and as an improved starting point for turbine anomaly detection within SHM. The NMSE is comparable to the standard GP but offers the extra security that the mean prediction is bounded, removing the potential for error that an unbounded model carries and providing the basis for increased operator confidence in the model.

\section{Conclusions} \label{Conclusion}

The novelty of this work was to propose and investigate two power curve models with physically meaningful predictions of mean and uncertainty. As discussed in ``Related work", this is a comparatively underdeveloped area of power curve modelling. Predictions of the mean and uncertainty that obey the physical laws imposed on the system being modelled are vital to understand in order to be able to extract meaningful information from these models. 

Of the two methods investigated, the authors have argued that HBP model is the most successful. When compared to the traditional GP, the HBP model performs favourably. The joint likelihood is significantly higher than the standard GP, showing that the bounded uncertainty region better captures the test points than its unbounded counterpart. This is an especially important result when the use of predictive uncertainty in further calculations is considered. The difference in NMSE between the standard GP and HBP is small enough to be functionally comparable, and ignores the fact that some of the predictions returned by the standard GP are not physically plausible. 

Predictions from the standard GP will not necessarily exceed the bounds with its mean predictions, however the model cannot provide confidence that its mean predictions are guaranteed to be physically plausible. The HBP is therefore superior in establishing operator trust and reliance on the power curve model, even when compared to a theoretically perfect standard GP. 

In conclusion, the HBP model is an important step in better quantifying the uncertainty of power curve models. It was shown that bounding the GP to respect physical limitations significantly improves uncertainty quantification, a result with interesting ramifications for SHM and other industrial use moving forward. 

\subsection{Further Work} \label{Further Work}

While the proposed methods for bounding the Gaussian Process met with varying levels of success, both models open up avenues for further work. It was clear that the heteroscedastic GP did not perform adequately in the warped space. Several methods to remedy this exists, and although outside the scope of this current work, may be considered in the future, including using Bayesian committees to tackle the truncated nature of the warped data shape or experimenting with alternative warping functions. 

The GP with Beta likelihood was successful, and while the fast approximation of the likelihood with the heteroscedastic method worked well, the random nature of the sampling and unification of the various distributions into an average distribution at each test point is not the most statistically robust method. Uniting the Beta distributions of each sample mathematically rather than using the mean and total variance would be more robust. Finally, it may be worth considering again the use of committee machines to better handle the transition to rated power seen in the power curve.

\section*{Acknowledgements}
The authors gratefully acknowledge the support of the Engineering Physical Sciences Research Council (EPSRC) through grant numbers EP/S001565/1 and EP/W002140/1. For the purpose of open access, the authors have applied a Creative Commons Attribution (CC-BY) licence to any Author Accepted Manuscript version arising.  

\bibliography{refs.bib}


\end{document}